\def\year{2024}\relax
\documentclass[letterpaper]{article} 
\usepackage{aaai22}  
\usepackage{times}  
\usepackage{helvet}  
\usepackage{courier}  
\usepackage[hyphens]{url}  
\usepackage{graphicx} 
\usepackage{natbib}  
\usepackage{caption} 
\DeclareCaptionStyle{ruled}{labelfont=normalfont,labelsep=colon,strut=off} 
\usepackage{algorithm}
\usepackage{algorithmic}

\usepackage{newfloat}
\usepackage{listings}
\floatstyle{ruled}
\newfloat{listing}{tb}{lst}{}
\floatname{listing}{Listing}

\usepackage{array,multirow,graphicx}
\newcolumntype{?}{!{\vrule width 0.5pt}}
\usepackage{rotating}
\usepackage{makecell}

\usepackage{xcolor}

\title{Are Large Language Models Good at Detecting Propaganda?}
\author{
    Julia Jose, Rachel Greenstadt
}
\affiliations{
    New York University\\

    Department of Computer Science and Engineering\\
    New York, New York, USA\\
    julia.jose@nyu.edu, greenstadt@nyu.edu
}

\begin{document}

\maketitle

\begin{abstract}
Propagandists use rhetorical devices that rely on logical fallacies and emotional appeals to advance their agendas. Recognizing these techniques is key to making informed decisions. Recent advances in Natural Language Processing (NLP) have enabled the development of systems capable of detecting manipulative content. In this study, we look at several Large Language Models and their performance in detecting propaganda techniques in news articles. We compare the performance of these LLMs with transformer-based models. We find that, while GPT-4 demonstrates superior F1 scores (F1=0.16) compared to GPT-3.5 and Claude 3 Opus, it does not outperform a RoBERTa-CRF baseline (F1=0.67). Additionally, we find that all three LLMs outperform a Multi-Granularity Network (MGN) baseline in detecting instances of one out of six propaganda techniques (name-calling), with GPT-3.5 and GPT-4 also outperforming the MGN baseline in detecting instances of appeal to fear and flag-waving.
\end{abstract}

\section{Introduction}
Propaganda is defined by Jowett \& O'Donnell (2006) as ``\textit{the deliberate, systematic attempt to shape perceptions, manipulate cognitions, and direct behavior to achieve a response that furthers the desired intent of the propagandist}''~\cite{jowett2018propaganda}. Manipulative information, as seen in the 2016 US presidential campaign, can harm society by influencing public opinion~\cite{brattberg2018russian, golovchenko2020cross}. Propaganda, however, is not a new concept and dates back to the Second World War. Scholars and educators around then emphasized the need to be able to identify propaganda~\cite{grahampropaganda,boothpropaganda,ernestpropaganda}. One of the most prominent detection strategies was to analyze pieces of information for the use of propaganda techniques and to ``guard against them"~\cite{grahampropaganda}. The Institute of Propaganda Analysis (IPA) was formed with this goal in mind. Outlined in their book \textit{The Fine Art of Propaganda: A Study of Father Coughlin's Speeches}~\cite{lee1939fine} are seven such propaganda techniques. Examples of some of these techniques include — 
\begin{itemize}
\item Name-Calling: ``Giving an idea a bad label and therefore rejecting and condemning it without examining the evidence.''
\item Bandwagon: ``Has as its theme `everybody - at least all of us - is doing it!' and thereby tries to convince the members of a group that their peers are accepting the program and that we should all jump on the bandwagon rather than be left out.''
\end{itemize}
 
Building on this work,  Da San Martino et al. (2019) derived eighteen such propaganda techniques that are widely used in news articles today.

Analyzing news articles to check for the presence of propaganda techniques~\cite{da2020prta} could help users gain a deeper understanding of the information's quality. To automate labeling these techniques in news articles at large scale, we aim to measure the effectiveness of Large Language Models (LLMs) on the task of detecting six out of the eighteen techniques mentioned in  Da San Martino et al. (2019). Specifically, we evaluate the following models -- OpenAI GPT-3.5, OpenAI GPT-4, and Anthropic Claude 3 Opus under different settings such as zero-shot, one-shot, chain-of-thought prompting~\cite{wei2022chain} and other advanced prompting techniques such as generated knowledge promoting~\cite{liu2021generated} and self-consistency prompting~\cite{wang2022self}. 

We find that none of the LLMs outperform a baseline model (a RoBERTa-CRF model with an ensemble of models for final classification layer~\cite{jurkiewicz2020applicaai} both at the macro-F1 level (average across all six techniques) and individual technique F1 scores. GPT-4 outperforms GPT-3.5 and Claude 3 Opus at the macro-F1 level.

We also compare the results to another baseline released by the dataset authors~\cite{martino2019fine}, which is a Multi-Granularity Network (MGN) model built on BERT embeddings. 
At the macro-F1 level, we find that GPT-3.5 and Claude 3 Opus do not outperform this MGN model. GPT-4 outperforms this model under one-shot, generated knowledge, and self-consistency prompting mechanisms. Looking at the techniques specifically, we find that all the LLMs outperform the MGN model in detecting instances of name-calling. GPT-4 and GPT-3.5 (only generated knowledge prompting) outperforms this model in detecting appeal-to-fear instances. Similarly, GPT-4 and GPT-3.5 (only one-shot) outperforms this model in detecting instances of flag-waving. None of the LLMs outperform the MGN model in detecting instances of loaded language, doubt, and exaggeration/minimization.

\begin{table*}[t]
\centering
  \begin{tabular}{l|p{0.73\linewidth}}
    \textbf{Propaganda Device}  & \textbf{Definition} \\[0.2cm] 
    \hline\\
    Name-Calling & ``Labeling the object of the propaganda campaign as either something the target audience fears, hates, finds undesirable or otherwise loves or praises'' \\[0.6cm] 
    Loaded Language & ``Using words or phrases with strong emotional implications to influence an audience'' \\[0.2cm] 
    Doubt & ```Questioning the credibility of someone or something'' \\[0.2cm] 
    Appeal to Fear & ``Seeking to build support for an idea by instilling anxiety and/or panic in the population towards an alternative, possibly based on preconceived judgments'' \\[0.6cm] 
    Flag-Waving & ``Playing on strong national feeling (or with respect to a group, e.g., race, gender, political preference) to justify or promote an action or idea'' \\[0.6cm] 
    Exaggeration or minimization & ``Either representing something in an excessive manner: making things larger, better, worse (e.g., “the best of the best”, “quality guaranteed”) or making something seem less important or smaller than it actually is '' \\[1cm] 
    \multicolumn{2}{p{0.9\linewidth}}{\small Note. From "Fine-Grained Analysis of Propaganda in News Articles", by Martino et al., 2019, EMNLP-IJCNLP, pp. 5636-5646~\cite{martino2019fine}}\\
  \end{tabular}
    \caption{Propaganda Techniques and Definitions used in this study}
  \label{table:propaganda devices}
\end{table*}

\section{Related Work}
\subsection{Language Model Evaluation}
Language Models have been evaluated on several domain-specific tasks. Wang et al.(2023) evaluated LLMs on natural language understanding in the health domain. They used clinical benchmark datasets and found that GPT-4 outperforms GPT-3.5 and Bard in text classification, named entity recognition, relation extraction, and natural language inference. Rehana et al. (2023) compared LLMs to BERT-based models on protein-protein interaction extraction and found that domain-specific models such as BioBERT and PubMedBERT outperform GPT-4. Li et al. (2023b) similarly evaluated LLMs in the finance domain and found that GPT-4 outperforms all other LLMs in news classification and sentiment analysis. Domain-specific models, however, outperformed GPT-4 in named entity recognition~\cite{li2023chatgpt}. Li et al. (2023a) evaluated LLMs in detecting instances of hate speech on social media under different prompt settings and found the results to be better than labels obtained from crowd-sourced workers. Examples where LLMs outperform state-of-the-art models include information extraction~\cite{wan2023gpt, ma2023large}, question answering~\cite{bang2023multitask}, machine translation~\cite{hendy2023good,wang2023document}, and information retrieval~\cite{ziems2023large}. 

\subsection{Prompt Engineering}
A prompt is a set of instructions describing the task that you want the LLM to accomplish. Prompts can be engineered to elicit more optimal responses. 

Zero-shot prompting refers to curating a prompt describing the task that you want the LLM to perform without giving it any examples pertaining to the task. Several studies have looked at using zero-shot prompts with GPT-3.5 and GPT-4 ~\cite{wei2023zero, espejel2023gpt}. Few-shot prompting, on the other hand, adds a bunch of examples to the task description to make instructions clearer~\cite {brown2020language}. Few-shot prompts have also been evaluated in tasks that require more complex reasoning from the LLMs such as in answering questions in radiation oncology physics~\cite{holmes2023evaluating}. Liu et al. (2021) introduced generated knowledge prompting - a technique by which you get the LLM to generate knowledge related to the task before making the prediction. They test it on a variety of datasets in the commonsense reasoning domain and show that it outperforms zero-shot and few-shot settings. Wei et al. (2022) introduced Chain-of-Thought (CoT) prompting, a technique in which the model is given intermediate reasoning steps to evoke the LLM's reasoning abilities. They test the prompting technique on math problems and show that it outperforms basic prompts. Wang et al. (2022) introduced self-consistency prompting in which the LLM is asked to generate multiple outputs and a majority voting technique help decide the final output i.e. consider the most ``consistent" answer. Their method outperforms commonsense reasoning benchmarks when compared to CoT prompts. 

In this study, we experiment with five of these prompting strategies - zero-shot, one-shot, chain-of-thought, generated knowledge, and self-consistency prompting.

\def\arraystretch{1.5}%
\begin{table*}
  \begin{tabular}{|p{1.4cm}?p{0.48cm}|p{0.45cm}|p{0.45cm}|p{0.45cm}|p{0.45cm}|p{0.42cm}?p{0.48cm}|p{0.45cm}|p{0.45cm}|p{0.45cm}|p{0.45cm}?p{0.48cm}|p{0.45cm}|p{0.45cm}|p{0.45cm}|p{0.45cm}|p{0.45cm}|}
    \hline
    \multirow{2}{*}{\textbf{Technique}} &
      \multicolumn{6}{c?}{\textbf{GPT-3.5}} &
      \multicolumn{5}{c?}{\textbf{GPT-4}} &
      \multicolumn{4}{c?}{\textbf{Claude 3 Opus}} &
      \multicolumn{1}{p{0.68cm}?}{\textbf{MGN}} & 
      \multicolumn{1}{p{1.4cm}?}{\textbf{\makecell[l]{RoBERTa\\-CRF}}} \\
    & \parbox{0.48cm}{Zero-Shot} & \parbox{0.45cm}{One-Shot} & \parbox{0.45cm}{CoT}  & \parbox{0.45cm}{GK}  & \parbox{0.45cm}{SC}  & \parbox{0.45cm}{FT} & \parbox{0.48cm}{Zero-Shot} & \parbox{0.5cm}{One-Shot} & \parbox{0.45cm}{CoT}  & \parbox{0.45cm}{GK}  & \parbox{0.45cm}{SC}  & \parbox{0.48cm}{Zero-Shot} & \parbox{0.45cm}{One-Shot} & \parbox{0.45cm}{CoT}  & \parbox{0.45cm}{GK} & & \\ [0.2cm]
    \hline
    Name-Calling & 0.21 & 0.13 & 0.13 & 0.25 & 0.24 & 0.23 &   0.27 & 0.30 & 0.18 & \underline{0.31} & 0.30   & 0.24 & 0.26 & 0.12  & 0.28  & 0 & \textbf{0.74}  \\
    \hline
    Loaded Language & 0.2 & 0.26 & 0.12 & 0.25 & 0.20 & 0.28 &   0.17 & 0.24 & 0.13 & 0.22 & 0.21  & 0.18 & 0.25 & 0.16  &0.28 & 0.40 & \textbf{0.80} \\
    \hline
    Doubt & 0.11 & 0.10 & 0.08 & 0.06 & 0.10 & 0.16 &    0.08 & 0.11 & 0.13 & 0.17 & 0.14    & 0.11 & 0.13 & 0.15  &0.10 & 0.19 & \textbf{0.63}  \\
    \hline
    Appeal to Fear & 0.04 & 0.07 & 0.04 & \underline{0.16} & 0.05 & 0.03 &   0.09 & 0.09 & 0.11 & 0.09 & 0.09    & 0.07 & 0.02 & 0.07  &0.03 & 0.09 & \textbf{0.48} \\
    \hline
   Flag-Waving & 0.07 & 0.09 & 0.06 & 0.01 & 0.07 &0 &    0.08 & 0.09 & \underline{0.14} & 0.11 & \underline{0.14}    & 0.04 & 0.05 & 0.07 &0.05 & 0.08 & \textbf{0.82}  \\
   \hline
    Exagger-\par ation or minimization & 0.09 & 0.11 & 0.05 & 0.09 & 0.09 & 0 &  0.07 & 0.10 & 0.08 & 0.09 & 0.09    & 0.08 & 0.06 & 0.07 &0.09  & 0.11  & \textbf{0.60} \\
   \hline
   \textbf{Macro-F1} & 0.12 & 0.13 & 0.08 & 0.14 & 0.12 & 0.11 &  0.13 & 0.15 & 0.13 & \underline{0.16} & \underline{0.16}   & 0.12 & 0.13 & 0.12 &0.14  & 0.14 &  \textbf{0.67} \\
   \hline
  \end{tabular}
  \caption{F1 scores of LLMs under different prompt settings across six propaganda techniques compared to the baseline model. (MGN=Multi-Granularity Network, FT = Fine-Tuned, CoT=Chain-of-Thought Prompting, GK=Generated Knowledge Prompting, SC=Self-Consistency)}
\label{table:technique f1}
\end{table*}

\section{Methods}

\subsection{Dataset}
We perform experiments on the PTC (Propaganda Techniques Corpus) dataset from Da San Martino et al. (2019). This dataset was developed by gathering propagandistic articles from sources listed under Media Bias/Fact Check. Da San Martino et al. (2019) partnered with media intelligence professionals to develop annotations of eighteen propaganda techniques used in these articles. The dataset contains articles with phrase-level instances of these techniques. For the scope of this study (concerning training/inference cost, speed, and time), we choose to focus on six out of eighteen of these techniques (as seen in Table~\ref{table:propaganda devices}) which are also some of the more common techniques used in news articles~\cite{martino2019fine}. We also made sure to focus on techniques that were more emotionally appealing (for example, by including ~\textit{flag-waving} instead of ~\textit{repetition}).

\subsection{Models}
We evaluate three large language models on the task of detecting propaganda techniques in news articles. The first is GPT-3.5 for which we use gpt-3.5-turbo-0125 in the OpenAI API. The second is GPT-4 for which we use gpt-4-0125-preview in the OpenAI API. The third is Claude 3 Opus for which we use claude-3-opus-20240229 in the Anthropic API. 

Propaganda techniques rely on emotional and logically flawed reasoning and hence we chose models that were found to be superior in text classification~\cite{wang2023large}, natural language understanding~\cite{zhong2023can}, reasoning~\cite{espejel2023gpt, li2023chatgpt} and entity extraction~\cite{ma2023large} among other benchmarks. We experiment with multiple prompt engineering techniques such as zero-shot, one-shot, chain-of-thought, generated knowledge, and self-consistency prompting. A table containing a list of the prompts used in our study can be found in the Appendix. 

\textbf{Fine-tuning GPT-3.5} We fine-tuned individual models for detecting each of the six techniques. We used the gpt-3.5-turbo-0125 model in the OpenAI API and used default hyperparameter values (epochs = 3, learning rate multiplier = 2, batch size = 1).

\textbf{Baseline} 
The authors of the PTC dataset organized a propaganda detection task at the International Workshop on Semantic Evaluation 2020 (SemEval 2020 Task 11) which received 44 submissions~\cite{martino2020semeval}. We compare our results to the highest-achieving model (by a team named ApplicaAI) for this task which used a RoBERTa-CRF model with an ensemble of models as the final classifier~\cite{jurkiewicz2020applicaai}. The ensemble models include a RoBERTa-CRF trained on the original data and also a model trained on additional data that was generated using the original RoBERTa-CRF model. We use this model as our upper bound to draw comparisons to the current state-of-the-art. 

We also compare our results to a model released by the dataset authors. This model is a novel Multi-Granularity Network (MGN) model that uses BERT embeddings and both sentence-level and phrase-level information while fine-tuning the model. Specifically, if the sentence is considered to be non-propagandistic then having the model not check for phrase-level instances of propaganda in it gives higher precision (hence higher F1 scores) compared to a BERT-based baseline fine-tuned to detect instances of these techniques in sentences~\cite{martino2019fine}. This model acts as our lower bound for evaluating the effectiveness of LLMs.

\section{Results}
We report model outcomes under different prompting strategies for all six techniques in Table~\ref{table:technique f1}. An extended version of the table including precision and recall values can be found in the Appendix. We compare these with the RoBERTa-CRF baseline and the Multi-Granularity Network (MGN) implementation.

\textbf{Name-Calling}: As seen in Table~\ref{table:technique f1}, none of the LLMs outperform the RoBERTa-CRF baseline (F1=0.74). While all versions of LLMs outperformed the MGN model, GPT-4 generated knowledge prompting gave us the highest F1 score of 0.31 with precision=0.34, recall=0.29 (see Table ~\ref{table:precision_recall_GPT4}). 

\textbf{Loaded-Language}: None of the LLMs outperform the RoBERTa-CRF baseline (F1=0.80). Furthermore, none of the LLMs outperformed the MGN model either (F1=0.40). 

\textbf{Doubt}: None of the LLMs outperform the RoBERTa-CRF baseline (F1=0.63). Furthermore, none of the LLMs outperformed the MGN model either (F1=0.19). 

\textbf{Appeal to Fear}: None of the LLMs outperformed the RoBERTa-CRF baseline (F1=0.48). GPT-3.5 generated knowledge prompting outperformed the MGN model and all other LLMs (with F1=0.16, precision=0.16, recall=0.15) but not the RoBERTa-CRF baseline. GPT-4 outperformed and/or was on par with the MGN model under all prompt settings. 

\textbf{Flag-Waving}: None of the LLMs outperformed the RoBERTa-CRF baseline (F1=0.82). All versions except zero-shot GPT-4 outperformed the MGN model, with GPT-4 self-consistency prompting giving us the highest F1 score of 0.144 (precision=0.13, recall=0.15). GPT-3.5 one-shot also outperformed the MGN model. 

\textbf{Exaggeration or Minimisation}: None of the LLMs outperformed the RoBERTa-CRF baseline (F1=0.60). GPT-3.5 one-shot performed closely with the MGN model, giving us an F1 score of 0.111 (precision=0.06, recall=0.52) whereas the MGN model gave an F1-score of 0.116. 

GPT-4 under one-shot, generated knowledge, and self-consistency prompting mechanisms outperform the MGN model (macro-F1). GPT-4 also outperforms GPT-3.5 and Claude 3 Opus (macro-F1). None of the LLMs outperform the RoBERTa-CRF baseline. 

\section{Discussions}
We explored the effectiveness of LLMs in detecting propaganda techniques in news articles. We find that none of these LLMs outperform a RoBERTa-CRF model with an ensemble of models in the final classification layer~\cite{jurkiewicz2020applicaai}. Chernyavskiy, Ilvovsky, and Nakov (2020) also used an ensemble of RoBERTa models for the propaganda detection task (SemEval 2020 Task 11). None of the LLMs outperform this model as well. Patil, Singh, and Agarwal (2020) used an ensemble of BERT and logistic regression along with features such as TF-IDF. This model also shows superior performance to the LLMs (Table~\ref{table:technique f1 expanded} in the Appendix contains F1 scores corresponding to these models). However, we find that GPT-3.5 and GPT-4 performs better than a Multi-Granularity Network based model~\cite{martino2019fine} in detecting three out of six techniques. We also find that GPT-4 outperforms the MGN model (macro-F1).

To strengthen the experiment's rigor, we experimented with multiple variations of the prompts for each setting. The F1 scores reported in Table~\ref{table:technique f1} correspond to prompts that gave us the highest macro-F1 across these variations. For example, for zero-shot prompting, we experimented with six different variations of prompts including the example prompt seen in~\cite{li2023hot} which gave us the highest macro-F1 score. Similarly, for one-shot prompting, we experimented with three variations. For chain-of-thought prompting, we experimented with three variations including the example seen in~\cite{kojima2022large}. Similarly, we tried two variations for self-consistency prompting and one variation for generated knowledge prompting. We believe this enhances the rigor of the conclusions drawn in this study.

We fine-tuned GPT-3.5 using a zero-shot prompt in the dataset's prompt template. When compared with their zero-shot counterpart (GPT-3.5 zero-shot in Table~\ref{table:technique f1}), the fine-tuned version gave better F1 scores for name-calling, loaded language, and doubt but worse scores for appeal to fear, flag-waving, and exaggeration/minimization. Future work could look into experimenting with the prompts used to fine-tune these models. Furthermore, our few-shot prompt consisted of one example (making it a one-shot prompt). Future studies could look into using more than one example while also being mindful of context window limitations. Researchers could also look into including examples that are related to the text at hand using information retrieval concepts based on semantic similarity~\cite{nashid2023retrieval}.

Da San Martino et al. (2019) report a moderate Inter Annotator Agreement (IAA) score between annotators while labeling the dataset. Their initial annotation stage (of the two-stage annotation process) saw a much lower IAA score of 0.24 and 0.28. They attribute this to cases where an annotator initially misses an instance but later agrees upon it in the second stage with a consolidator. Looking at the precision values in Table~\ref{table:precision_recall_GPT3.5}, ~\ref{table:precision_recall_GPT4}, ~\ref{table:precision_recall_claude3}, we see several instances with really low precision values (indicating high false positives). This raises the question of whether the LLMs are identifying instances of propaganda techniques that were missed by human annotators in the original dataset. Further evaluation of the quality of these labels might help comprehend this.

\section{Conclusions}
While LLMs have achieved state-of-the-art results at numerous tasks such as information extraction~\cite{wan2023gpt, ma2023large}, question answering~\cite{bang2023multitask}, machine translation~\cite{hendy2023good,wang2023document}, and information retrieval~\cite{ziems2023large}, we find that LLMs perform inadequately at detecting propaganda techniques in news articles on the PTC dataset when compared to a RoBERTa-CRF baseline. A fine-tuned version of GPT-3.5 also gave us sub-par results. We find that GPT-4 outperforms a Multi-Granularity Network (MGN) baseline as well as GPT-3.5 and Claude 3 Opus (macro-F1). We also find that GPT-3.5 and GPT-4 outperforms the MGN model in detecting instances of name-calling, appeal to fear, and flag-waving, with Claude 3 Opus outperforming detection on only one technique (name-calling).

\section*{Acknowledgements} This work was supported by the National Science Foundation under grant number 1940713.

\appendix
\section{Appendix}
\begin{table*}[t]
\centering
  \begin{tabular}{p{2cm}|p{0.8\linewidth}}
    \textbf{Prompt Setting}  & \textbf{Prompt} \\
    \hline\\
    Zero-shot & I want you to perform a data annotation task. In your output, I want you to return a json dictionary with key as phrase and value as technique, depending on whether you think the phrases in the following text uses $<technique>$. 
A phrase is $<technique>$ if you perceive that it is $<technique\ definition>$. 
I want you to respond with a json dictionary strictly having the detected phrases as keys and technique ($<technique>$) as value (eg, {``phrase'':``$<technique>$'',...}) and provide an explanation for your output in the end. 

Text:  
\\

One-shot & I want you to perform a data annotation task. In your output, I want you to return a json dictionary with key as phrase and value as technique, depending on whether you think the phrases in the following text uses $<technique>$. 
A phrase is $<technique>$ if you perceive that it is $<technique\ definition>$. 
I want you to respond with a json dictionary strictly having the detected phrases as keys and technique ($<technique>$) as value (eg, {``phrase'':``$<technique>$'',...}) and provide an explanation for your output in the end.

Text:
$<example\ article>$

Output:

\{
``$<phrase 1>$": "$<technique>$''
\}

Explanation:

- $<phrase 1>$ and $<explanation>$

Text: 
 \\

CoT & Using the definition of the technique given to you below, identify and extract instances of the technique in the text below by strictly following this format:

(1) For each phrase in the text, return the following in your answer: 

- $<phrase>$ Let's think step by step. $<explanation>$ $<yes$ or $no>$

(2) Finally return a json dictionary strictly making sure that each detected phrase (from step 1) is a key and the technique ($<technique>$) is its value (eg, {``phrase'':``$<technique>$'',...}). 

Technique: 

``$<technique>$'': ``$<technique\ definition>$''.

Text:
 \\

Generated Knowledge Prompting & Detect the propaganda technique $<technique>$ in the input article. Follow the example format given below: 

Input: Article: $<example\ article>$

are there instances of $<technique>$ in this text?

Knowledge: $<technique>$ is the $<technique\ definition>$. The phrase ``$<phrase\ 1>$'' is an instance of $<technique>$ because it $<explanation>$. 

Output:

\{
``$<phrase 1>$": "$<technique>$''
\}

Input: Article: $<article>$ 

are there instances of $<technique>$ in this text?

Knowledge: 
 \\

Self-Consistency & I want you to perform a data annotation task. In your output, I want you to return a json dictionary with key as phrase and value as technique, depending on whether you think the phrases in the following text uses doubt. 
A phrase is $<technique>$ if you perceive that it is $<technique\ definition>$
I want you to respond with a json dictionary strictly having the detected phrases as keys and technique ($<technique>$) as value. For each detection, explain your reasoning at the very beginning before the final dictionary output.

An example is given below:  

Text:

$<example\ article>$

Explanation:

- $<phrase 1>$ and $<explanation>$

Output:

\{
``$<phrase 1>$": "$<technique>$''
\}

Text: $<article>$ 

Explanation:
  \end{tabular}
    \caption{Prompts used in this study}
  \label{table:prompts}
\end{table*}


\def\arraystretch{1.5}%
\begin{sidewaystable}
\centering
  \begin{tabular}{|p{2cm}?l|l|l|l|l|l?l|l|l?l|l|l?l|l|l?l|l|l|}
    \hline
    \multirow{2}{*}{\textbf{Technique}} &
      \multicolumn{3}{c?}{\textbf{Zero-Shot}} &
      \multicolumn{3}{c?}{\textbf{One-Shot}} &
      \multicolumn{3}{c?}{\textbf{CoT}} &
      \multicolumn{3}{c?}{\textbf{GK}} &
      \multicolumn{3}{c?}{\textbf{SC}} &
      \multicolumn{3}{c|}{\textbf{FT}} \\
    & \parbox{0.5cm}{F1} & \parbox{0.5cm}{P} & \parbox{0.5cm}{R}& \parbox{0.5cm}{F1} & \parbox{0.5cm}{P} & \parbox{0.5cm}{R}& \parbox{0.5cm}{F1} & \parbox{0.5cm}{P} & \parbox{0.5cm}{R}& \parbox{0.5cm}{F1} & \parbox{0.5cm}{P} & \parbox{0.5cm}{R}& \parbox{0.5cm}{F1} & \parbox{0.5cm}{P} & \parbox{0.5cm}{R} & \parbox{0.5cm}{F1} & \parbox{0.5cm}{P} & \parbox{0.5cm}{R}  \\ [0.2cm]
    \hline
    Name-Calling & 0.21 & 0.18 & 0.25 & 0.13 & 0.18 & 0.10 & 0.13 & 0.10 & 0.17& 0.25 &  0.20 &  0.32& 0.24&0.23&0.25& 0.239& 0.298&  0.19 \\
    \hline
    Loaded Language & 0.2 & 0.15 & 0.29 & 0.26 & 0.20 & 0.37 & 0.12 &0.10 &0.15& 0.25 & 0.19& 0.35& 0.20&0.19&0.21& 0.287& 0.289& 0.28 \\
    \hline
    Doubt & 0.11 & 0.06 & 0.32 & 0.10 & 0.06 & 0.34 & 0.08 &0.05 &0.24& 0.06&  0.12&  0.04& 0.10&0.08&0.13& 0.167& 0.215& 0.13 \\
    \hline
    Appeal to Fear & 0.04 & 0.02 & 0.14 & 0.07 & 0.04 & 0.27 & 0.04 &0.024  &0.13& 0.161&  0.16&  0.15& 0.05&0.03&0.10&0.038& 0.138& 0.02 \\
    \hline
   Flag-Waving & 0.07 & 0.04 & 0.19 & 0.09 & 0.06 & 0.19 & 0.06 &0.04 &0.13& 0.01& 0.03& 0.01&0.07&0.07&0.08 & 0& 0& 0\\
   \hline
    Exaggeration or minimization & 0.09 & 0.05 & 0.46 & 0.11 & 0.06 & 0.52 & 0.05 &0.03 &0.22& 0.09& 0.05& 0.33& 0.09&0.05&0.27 & 0 & 0& 0\\
   \hline
  \end{tabular}

  \caption{GPT-3.5 Precision, Recall, F1 scores (FT = Fine-Tuned, CoT=Chain-of-Thought Prompting, GK=Generated Knowledge Prompting, SC=Self-Consistency)}
\label{table:precision_recall_GPT3.5}
\end{sidewaystable}


\def\arraystretch{1.5}%
\begin{sidewaystable}
\centering

  \begin{tabular}{|p{2cm}?l|l|l|l|l|l?l|l|l?l|l|l?l|l|l|}
    \hline
    \multirow{2}{*}{\textbf{Technique}} &
      \multicolumn{3}{c?}{\textbf{Zero-Shot}} &
      \multicolumn{3}{c?}{\textbf{One-Shot}} &
      \multicolumn{3}{c?}{\textbf{CoT}} &
      \multicolumn{3}{c?}{\textbf{GK}} &
      \multicolumn{3}{c?}{\textbf{SC}} \\
    & \parbox{0.5cm}{F1} & \parbox{0.5cm}{P} & \parbox{0.5cm}{R}& \parbox{0.5cm}{F1} & \parbox{0.5cm}{P} & \parbox{0.5cm}{R}& \parbox{0.5cm}{F1} & \parbox{0.5cm}{P} & \parbox{0.5cm}{R}& \parbox{0.5cm}{F1} & \parbox{0.5cm}{P} & \parbox{0.5cm}{R}& \parbox{0.5cm}{F1} & \parbox{0.5cm}{P} & \parbox{0.5cm}{R}\\ [0.2cm]
    \hline
    Name-Calling &0.274& 0.213& 0.385 & 0.307&0.243&0.416 &0.184&0.155&0.227&0.316&0.347&0.29&0.309&0.369&0.266\\
    \hline
    Loaded Language & 0.173& 0.111& 0.389 & 0.24&0.174&0.384&0.13&0.099&0.186&0.225&0.147&0.476&0.21&0.184&0.244\\
    \hline
    Doubt & 0.082& 0.045& 0.464 &0.115&0.066&0.42&0.135&0.08&0.425&0.172&0.116&0.33&0.144&0.093&0.324\\
    \hline
    Appeal to Fear & 0.096& 0.054& 0.427& 0.093&0.054&0.341 &0.118&0.073&0.299&0.094&0.062&0.195&0.096&0.071&0.147\\
    \hline
   Flag-Waving &0.08& 0.052& 0.177& 0.09&0.057&0.205&0.142&0.153&0.132&0.116&0.113&0.12&0.144&0.136&0.154
\\
   \hline
    Exaggeration or minimization & 0.079& 0.043& 0.526& 0.104&0.057&0.554&0.087&0.05&0.355&0.09&0.059&0.19&0.096&0.057&0.3\\
   \hline
  \end{tabular}

  \caption{GPT-4 Precision, Recall, F1 scores (CoT=Chain-of-Thought Prompting, GK=Generated Knowledge Prompting, SC=Self-Consistency)}
\label{table:precision_recall_GPT4}
\end{sidewaystable}


\def\arraystretch{2}%
\begin{table*}
  \begin{tabular}{|p{2.5cm}?l|l|l|l|l|l?l|l|l?l|l|l|}
    \hline
    \multirow{2}{*}{\textbf{Technique}} &
      \multicolumn{3}{c?}{\textbf{Zero-Shot}} &
      \multicolumn{3}{c?}{\textbf{One-Shot}} &
      \multicolumn{3}{c?}{\textbf{CoT}} &
      \multicolumn{3}{c?}{\textbf{GK}} \\
    & \parbox{0.5cm}{F1} & \parbox{0.5cm}{P} & \parbox{0.5cm}{R}& \parbox{0.5cm}{F1} & \parbox{0.5cm}{P} & \parbox{0.5cm}{R}& \parbox{0.5cm}{F1} & \parbox{0.5cm}{P} & \parbox{0.5cm}{R}& \parbox{0.5cm}{F1} & \parbox{0.5cm}{P} & \parbox{0.5cm}{R}\\ [0.2cm]
    \hline
    Name-Calling &0.245&0.229&0.264& 0.267&0.307&0.236&0.222&0.243&0.205&0.283&0.317&0.255\\
    \hline
    Loaded Language & 0.184&0.153&0.229&0.259&0.217&0.322 &0.166&0.164&0.169&0.288&0.235&0.373\\
    \hline
    Doubt & 0.113&0.074& 0.246& 0.132&0.11&0.166&0.153&0.103&0.296&0.102&0.25&0.064\\
    \hline
    Appeal to Fear & 0.075&0.046&0.194 & 0.027&0.02&0.043&0.075&0.052&0.134&0.033&0.034&0.033\\
    \hline
   Flag-Waving & 0.044&0.032&0.069& 0.052&0.048&0.057&0.073&0.059&0.094&0.058&0.102&0.041\\
   \hline
    Exaggeration or minimization & 0.082&0.047&0.292&0.069&0.041&0.203&0.073&0.048&0.155&0.098&0.066&0.183\\
   \hline
  \end{tabular}

  \caption{Claude 3 Opus Precision, Recall, F1 scores (CoT=Chain-of-Thought Prompting, GK=Generated Knowledge Prompting)}
\label{table:precision_recall_claude3}
\end{table*}

\def\arraystretch{2}%
\begin{table*}
\centering
  \begin{tabular}{|p{3cm}?p{3.5cm}|p{2.5cm}|}
    \hline
    \multirow{2}{*}{\textbf{Technique}} &
      \multicolumn{1}{c?}{\textbf{\thead{Chernyavskiy,\\ Ilvovsky, and Nakov (2020)}}} &
      \multicolumn{1}{c?}{\textbf{\thead{Patil, Singh,\\ and Agarwal (2020) }}}  \\
    & \makecell[c]{F1} & \makecell[c]{F1}\\ [0.2cm]
    \hline
    Name-Calling & \makecell[c]{0.73} & \makecell[c]{0.70}\\
    \hline
    Loaded Language & \makecell[c]{0.80} & \makecell[c]{0.75} \\
    \hline
    Doubt & \makecell[c]{0.66} & \makecell[c]{0.52} \\
    \hline
    Appeal to Fear & \makecell[c]{0.45} & \makecell[c]{0.32} \\
    \hline
   Flag-Waving & \makecell[c]{0.74} & \makecell[c]{0.75} \\
   \hline
    Exaggeration or minimization & \makecell[c]{0.59} & \makecell[c]{0.49} \\
   \hline
   \textbf{Macro-F1} & \makecell[c]{0.66} & \makecell[c]{0.58} \\
   \hline
  \end{tabular}
  \caption{F1 scores of transformer-based models submitted to SemEval Task 11 2020}
\label{table:technique f1 expanded}
\end{table*}





\begin{thebibliography}{30}
\providecommand{\natexlab}[1]{#1}

\bibitem[{Bang et~al.(2023)Bang, Cahyawijaya, Lee, Dai, Su, Wilie, Lovenia, Ji,
  Yu, Chung et~al.}]{bang2023multitask}
Bang, Y.; Cahyawijaya, S.; Lee, N.; Dai, W.; Su, D.; Wilie, B.; Lovenia, H.;
  Ji, Z.; Yu, T.; Chung, W.; et~al. 2023.
\newblock A multitask, multilingual, multimodal evaluation of chatgpt on
  reasoning, hallucination, and interactivity.
\newblock \emph{arXiv preprint arXiv:2302.04023}.

\bibitem[{Booth;(1940)}]{boothpropaganda}
Booth;, G.~C. 1940.
\newblock Can Propaganda Analysis Be Taught?
\newblock \emph{Junior College Journal}, 310--312.

\bibitem[{Brattberg and Maurer(2018)}]{brattberg2018russian}
Brattberg, E.; and Maurer, T. 2018.
\newblock \emph{Russian election interference: Europe's counter to fake news
  and cyber attacks}, volume~23.
\newblock Carnegie Endowment for International Peace Washington, DC.

\bibitem[{Brown et~al.(2020)Brown, Mann, Ryder, Subbiah, Kaplan, Dhariwal,
  Neelakantan, Shyam, Sastry, Askell et~al.}]{brown2020language}
Brown, T.; Mann, B.; Ryder, N.; Subbiah, M.; Kaplan, J.~D.; Dhariwal, P.;
  Neelakantan, A.; Shyam, P.; Sastry, G.; Askell, A.; et~al. 2020.
\newblock Language models are few-shot learners.
\newblock \emph{Advances in neural information processing systems}, 33:
  1877--1901.

\bibitem[{Da~San~Martino et~al.(2019)Da~San~Martino, Seunghak,
  Barr{\'o}n-Cedeno, Petrov, Nakov et~al.}]{martino2019fine}
Da~San~Martino, G.; Seunghak, Y.; Barr{\'o}n-Cedeno, A.; Petrov, R.; Nakov, P.;
  et~al. 2019.
\newblock Fine-grained analysis of propaganda in news article.
\newblock In \emph{Proceedings of the 2019 conference on empirical methods in
  natural language processing and the 9th international joint conference on
  natural language processing (EMNLP-IJCNLP)}, 5636--5646. Association for
  Computational Linguistics.

\bibitem[{Da~San~Martino et~al.(2020)Da~San~Martino, Shaar, Zhang, Sh,
  Barr{\'o}n-Cedeno, Nakov et~al.}]{da2020prta}
Da~San~Martino, G.; Shaar, S.; Zhang, Y.; Sh, Y.; Barr{\'o}n-Cedeno, A.; Nakov,
  P.; et~al. 2020.
\newblock Prta: A system to support the analysis of propaganda techniques in
  the news.
\newblock In \emph{Proceedings of the 58th Annual Meeting of the Association
  for Computational Linguistics: System Demonstrations}, 287--293. ASSOC
  COMPUTATIONAL LINGUISTICS-ACL.

\bibitem[{Espejel et~al.(2023)Espejel, Ettifouri, Alassan, Chouham, and
  Dahhane}]{espejel2023gpt}
Espejel, J.~L.; Ettifouri, E.~H.; Alassan, M. S.~Y.; Chouham, E.~M.; and
  Dahhane, W. 2023.
\newblock GPT-3.5, GPT-4, or BARD? Evaluating LLMs reasoning ability in
  zero-shot setting and performance boosting through prompts.
\newblock \emph{Natural Language Processing Journal}, 5: 100032.

\bibitem[{Golovchenko et~al.(2020)Golovchenko, Buntain, Eady, Brown, and
  Tucker}]{golovchenko2020cross}
Golovchenko, Y.; Buntain, C.; Eady, G.; Brown, M.~A.; and Tucker, J.~A. 2020.
\newblock Cross-platform state propaganda: Russian trolls on twitter and
  YouTube during the 2016 US Presidential Election.
\newblock \emph{The International Journal of Press/Politics}, 25(3): 357--389.

\bibitem[{Graham(1939)}]{grahampropaganda}
Graham, M. M.~W. 1939.
\newblock Analyzing Propaganda.
\newblock \emph{Proceedings of the National Education Association}, 423--31.

\bibitem[{Hendy et~al.(2023)Hendy, Abdelrehim, Sharaf, Raunak, Gabr,
  Matsushita, Kim, Afify, and Awadalla}]{hendy2023good}
Hendy, A.; Abdelrehim, M.; Sharaf, A.; Raunak, V.; Gabr, M.; Matsushita, H.;
  Kim, Y.~J.; Afify, M.; and Awadalla, H.~H. 2023.
\newblock How good are gpt models at machine translation? a comprehensive
  evaluation.
\newblock \emph{arXiv preprint arXiv:2302.09210}.

\bibitem[{Hollis;(1939)}]{ernestpropaganda}
Hollis;, E.~V. 1939.
\newblock Antidote for Propaganda,.
\newblock \emph{School and Society}, 50:449--453.

\bibitem[{Holmes et~al.(2023)Holmes, Liu, Zhang, Ding, Sio, McGee, Ashman, Li,
  Liu, Shen et~al.}]{holmes2023evaluating}
Holmes, J.; Liu, Z.; Zhang, L.; Ding, Y.; Sio, T.~T.; McGee, L.~A.; Ashman,
  J.~B.; Li, X.; Liu, T.; Shen, J.; et~al. 2023.
\newblock Evaluating large language models on a highly-specialized topic,
  radiation oncology physics.
\newblock \emph{Frontiers in Oncology}, 13.

\bibitem[{Jowett and O'Donnell(2018)}]{jowett2018propaganda}
Jowett, G.~S.; and O'Donnell, V. 2018.
\newblock \emph{Propaganda \& Persuasion}.
\newblock Sage publications.

\bibitem[{Jurkiewicz et~al.(2020)Jurkiewicz, Borchmann, Kosmala, and
  Grali{\'n}ski}]{jurkiewicz2020applicaai}
Jurkiewicz, D.; Borchmann, {\L}.; Kosmala, I.; and Grali{\'n}ski, F. 2020.
\newblock ApplicaAI at SemEval-2020 task 11: On RoBERTa-CRF, span CLS and
  whether self-training helps them.
\newblock \emph{arXiv preprint arXiv:2005.07934}.

\bibitem[{Kojima et~al.(2022)Kojima, Gu, Reid, Matsuo, and
  Iwasawa}]{kojima2022large}
Kojima, T.; Gu, S.~S.; Reid, M.; Matsuo, Y.; and Iwasawa, Y. 2022.
\newblock Large language models are zero-shot reasoners.
\newblock \emph{Advances in neural information processing systems}, 35:
  22199--22213.

\bibitem[{Lee and Lee(1939)}]{lee1939fine}
Lee, A.; and Lee, E.~B. 1939.
\newblock The fine art of propaganda.

\bibitem[{Li et~al.(2023{\natexlab{a}})Li, Fan, Atreja, and
  Hemphill}]{li2023hot}
Li, L.; Fan, L.; Atreja, S.; and Hemphill, L. 2023{\natexlab{a}}.
\newblock “HOT” ChatGPT: The promise of ChatGPT in detecting and
  discriminating hateful, offensive, and toxic comments on social media.
\newblock \emph{ACM Transactions on the Web}.

\bibitem[{Li et~al.(2023{\natexlab{b}})Li, Zhu, Ma, Liu, and
  Shah}]{li2023chatgpt}
Li, X.; Zhu, X.; Ma, Z.; Liu, X.; and Shah, S. 2023{\natexlab{b}}.
\newblock Are chatgpt and gpt-4 general-purpose solvers for financial text
  analytics? an examination on several typical tasks.
\newblock \emph{arXiv preprint arXiv:2305.05862}.

\bibitem[{Liu et~al.(2021)Liu, Liu, Lu, Welleck, West, Bras, Choi, and
  Hajishirzi}]{liu2021generated}
Liu, J.; Liu, A.; Lu, X.; Welleck, S.; West, P.; Bras, R.~L.; Choi, Y.; and
  Hajishirzi, H. 2021.
\newblock Generated knowledge prompting for commonsense reasoning.
\newblock \emph{arXiv preprint arXiv:2110.08387}.

\bibitem[{Ma et~al.(2023)Ma, Cao, Hong, and Sun}]{ma2023large}
Ma, Y.; Cao, Y.; Hong, Y.; and Sun, A. 2023.
\newblock Large language model is not a good few-shot information extractor,
  but a good reranker for hard samples!
\newblock \emph{arXiv preprint arXiv:2303.08559}.

\bibitem[{Martino et~al.(2020)Martino, Barr{\'o}n-Cedeno, Wachsmuth, Petrov,
  and Nakov}]{martino2020semeval}
Martino, G.; Barr{\'o}n-Cedeno, A.; Wachsmuth, H.; Petrov, R.; and Nakov, P.
  2020.
\newblock SemEval-2020 task 11: Detection of propaganda techniques in news
  articles.
\newblock \emph{arXiv preprint arXiv:2009.02696}.

\bibitem[{Nashid, Sintaha, and Mesbah(2023)}]{nashid2023retrieval}
Nashid, N.; Sintaha, M.; and Mesbah, A. 2023.
\newblock Retrieval-based prompt selection for code-related few-shot learning.
\newblock In \emph{2023 IEEE/ACM 45th International Conference on Software
  Engineering (ICSE)}, 2450--2462. IEEE.

\bibitem[{Wan et~al.(2023)Wan, Cheng, Mao, Liu, Song, Li, and
  Kurohashi}]{wan2023gpt}
Wan, Z.; Cheng, F.; Mao, Z.; Liu, Q.; Song, H.; Li, J.; and Kurohashi, S. 2023.
\newblock Gpt-re: In-context learning for relation extraction using large
  language models.
\newblock \emph{arXiv preprint arXiv:2305.02105}.

\bibitem[{Wang et~al.(2023)Wang, Lyu, Ji, Zhang, Yu, Shi, and
  Tu}]{wang2023document}
Wang, L.; Lyu, C.; Ji, T.; Zhang, Z.; Yu, D.; Shi, S.; and Tu, Z. 2023.
\newblock Document-level machine translation with large language models.
\newblock \emph{arXiv preprint arXiv:2304.02210}.

\bibitem[{Wang et~al.(2022)Wang, Wei, Schuurmans, Le, Chi, Narang, Chowdhery,
  and Zhou}]{wang2022self}
Wang, X.; Wei, J.; Schuurmans, D.; Le, Q.; Chi, E.; Narang, S.; Chowdhery, A.;
  and Zhou, D. 2022.
\newblock Self-consistency improves chain of thought reasoning in language
  models.
\newblock \emph{arXiv preprint arXiv:2203.11171}.

\bibitem[{Wang, Zhao, and Petzold(2023)}]{wang2023large}
Wang, Y.; Zhao, Y.; and Petzold, L. 2023.
\newblock Are large language models ready for healthcare? a comparative study
  on clinical language understanding.
\newblock In \emph{Machine Learning for Healthcare Conference}, 804--823. PMLR.

\bibitem[{Wei et~al.(2022)Wei, Wang, Schuurmans, Bosma, Xia, Chi, Le, Zhou
  et~al.}]{wei2022chain}
Wei, J.; Wang, X.; Schuurmans, D.; Bosma, M.; Xia, F.; Chi, E.; Le, Q.~V.;
  Zhou, D.; et~al. 2022.
\newblock Chain-of-thought prompting elicits reasoning in large language
  models.
\newblock \emph{Advances in neural information processing systems}, 35:
  24824--24837.

\bibitem[{Wei et~al.(2023)Wei, Cui, Cheng, Wang, Zhang, Huang, Xie, Xu, Chen,
  Zhang et~al.}]{wei2023zero}
Wei, X.; Cui, X.; Cheng, N.; Wang, X.; Zhang, X.; Huang, S.; Xie, P.; Xu, J.;
  Chen, Y.; Zhang, M.; et~al. 2023.
\newblock Zero-shot information extraction via chatting with chatgpt.
\newblock \emph{arXiv preprint arXiv:2302.10205}.

\bibitem[{Zhong et~al.(2023)Zhong, Ding, Liu, Du, and Tao}]{zhong2023can}
Zhong, Q.; Ding, L.; Liu, J.; Du, B.; and Tao, D. 2023.
\newblock Can chatgpt understand too? a comparative study on chatgpt and
  fine-tuned bert.
\newblock \emph{arXiv preprint arXiv:2302.10198}.

\bibitem[{Ziems et~al.(2023)Ziems, Yu, Zhang, and Jiang}]{ziems2023large}
Ziems, N.; Yu, W.; Zhang, Z.; and Jiang, M. 2023.
\newblock Large language models are built-in autoregressive search engines.
\newblock \emph{arXiv preprint arXiv:2305.09612}.

\end{thebibliography}
\end{document}